\documentclass[runningheads]{llncs}

\usepackage[T1]{fontenc}
\usepackage{graphicx}
\usepackage{subcaption}

\usepackage{hyperref}

\let\oldvec\vec
\usepackage{amsmath}
\renewcommand{\vec}[1]{\oldvec{#1}}

\usepackage{makecell}
\usepackage{adjustbox}
\usepackage{xcolor}
\usepackage{hyperref}
\usepackage{comment}
\usepackage[moderate]{savetrees} 

\usepackage{enumitem}
\newlist{questions}{enumerate}{2}
\setlist[questions,1]{label=RQ\arabic*.,ref=RQ\arabic*}
\setlist[questions,2]{label=(\alph*),ref=\thequestionsi(\alph*)}

\usepackage[most]{tcolorbox}

\begin{document}
\title{Leveraging Machine Learning and Enhanced
Parallelism Detection for BPMN Model
Generation from Text}
\titlerunning{BPMN Model Generation from Text}
%
\author{Phuong Nam Lê \and
Charlotte Schneider-Depré \and
Alexandre Goossens \and 
Alexander Stevens \and
Aurélie Leribaux \and
Johannes De Smedt}

\authorrunning{Phuong et al.}
%
\institute{Research Centre for Information Systems Engineering, KU Leuven, Leuven, Belgium}
\maketitle              
\begin{abstract}
Efficient planning, resource management, and consistent operations often rely on converting textual process documents into formal Business Process Model and Notation (BPMN) models. However, this conversion process remains time-intensive and costly. Existing approaches, whether rule-based or machine-learning-based, still struggle with writing styles and often fail to identify parallel structures in process descriptions.

This paper introduces an automated pipeline for extracting BPMN models from text, leveraging the use of machine learning and large language models. A key contribution of this work is the introduction of a newly annotated dataset, which significantly enhances the training process. Specifically, we augment the PET dataset with 15 newly annotated documents containing 32 parallel gateways for model training, a critical feature often overlooked in existing datasets. 
This addition enables models to better capture parallel structures, a common but complex aspect of process descriptions. The proposed approach demonstrates adequate performance in terms of reconstruction accuracy, offering a promising foundation for organizations to accelerate BPMN model creation.

\end{abstract}

\section{Introduction}

Organizations are increasingly focused on strategies to enhance planning clarity, resource alignment, and operational consistency. Business Process Management (BPM) plays a pivotal role in achieving these objectives by offering a systematic approach to analyze, optimize, and monitor business processes. Central to BPM is the use of formal models, which are essential for effective process implementation, communication, and automation \cite{friedrich2010automated}. The Business Process Model and Notation (BPMN) has become the de facto standard for modeling procedural workflows, originally developed by the Business Process Management Initiative (BPMI) and later standardized as BPMN 2.0 by the Object Management Group (OMG) \cite{BPMN_Spec}. 

Despite its widespread adoption, manually creating BPMN diagrams remains labor-intensive, requiring significant expertise often leading to slow and error-prone results \cite{Neuberger}. A major challenge in speeding up this process is that textual documents describing these business processes are ambiguous and contain diverse writing styles, which are difficult to automate. An additional challenge in process modeling is accurately representing parallelism—scenarios where multiple activities co-occur. Despite its critical role in capturing business process complexity, parallelism has received limited research attention. Existing methodologies primarily rely on rule-based and template-driven approaches \cite{FriedrichBPMN}, which struggle to adapt across different domains and writing styles \cite{Bellan_InContext2022}. 
\cite{bellan2021process} further highlight the need for larger and more diverse datasets to enable the development of better models to extract complex flow structures.

To address these limitations, this paper introduces a novel approach to automating BPMN model generation using Natural Language Processing (NLP) techniques. Specifically, it leverages pre-trained language models, such as Bidirectional Encoder Representations from Transformers (BERT) \cite{BERT} and the Robustly Optimized BERT Approach (RoBERTa) \cite{RoBERTa}, to extract and structure BPMN elements from textual descriptions. 
BERT-based models are relevant because their bidirectional contextual understanding, pre-trained language representations, and efficiency in token-level classification make them superior to traditional decoder models for tasks like NER \cite{RoBERTa}. 
While recently Large Language Model (LLM)-based approaches have surfaced which return strong performance through prompting and fine-tuning, including \cite{neuberger2024universal,grohs2023large,kourani2024process}, none have addressed parallelism explicitly.
\cite{kourani2024process} uses Partially Ordered Workflow Language (POWL) to ensure behavioral consistency, yet, a parallelism-specific evaluation is not performed.
Additionally, our contribution lies in enhancing the PET dataset \cite{PET} with process descriptions aimed at additional parallel behavior requiring \textit{AND Gateways}. 
This allows for a better ground truth evaluation of BPMN discovery techniques, including LLM versions which can be fine-tuned to our examples.

The remainder of this paper is structured as follows. Section \ref{sec: background} reviews existing BPMN generation methods, discusses state-of-the-art NLP techniques, and identifies research gaps. Section \ref{sec: Methodology} details the design of the proposed pipeline followed by the experiments in Section \ref{sec: evaluation}. Section \ref{sec: Discussion} interprets the experimental results and addresses the research questions. Finally, Section \ref{sec: Conclusion} summarizes the contributions of this research, discusses its implications, and suggests directions for future work.

\section{Background and Related Work}\label{sec: background}




Transforming textual descriptions into BPMN models encompasses several NLP sub-tasks \cite{Maqbool2019}. An example of an annotated process description of a library request reads \textit{``When a request for a book comes in, the library staff member consults the digital catalog to check for the book's availability. If the book is currently on loan or not in the library's collection, the staff member informs the requester right away. If the book is available, the staff member starts the checkout procedure by logging the book against the requester's library account \textit{\textbf{and simultaneously}} retrieving the book using the automated system.''}



Many NLP-based extraction techniques for process models perform a similar pipeline. Initially, the text undergoes pre-processing to clean, structure, and prepare the raw data for analysis, such as tokenization and part-of-speech tagging. 
Subsequently, Information Extraction (IE), a subfield of NLP, is employed to derive explicit semantic structures from the text \cite{Neuberger}. These tasks include identifying and categorizing entities, their relationships, and resolving coreferences \cite{GrishmanInfoExtraction}. All extracted elements are then organized to facilitate mapping into BPMN models \cite{GrishmanInfoExtraction}. 

\textbf{Tokenization} is an NLP technique that breaks down text into smaller, manageable units for analysis. Tokenization has been widely used in previous studies, primarily to parse and understand the syntactic structures of text segments \cite{friedrich2010automated,Neuberger,PET,FriedrichBPMN}, albeit not for BPMN models specifically.

\textbf{Part-Of-Speech (POS)} tagging assigns grammatical roles to each word, such as verb, noun, or adjective \cite{Neuberger}. This step aids in understanding sentence structure, which is beneficial for tasks like Named Entity Recognition (NER) and Relation Extraction (RE). For example, research in \cite{Neuberger} illustrates the use of POS tags to inform the training parameters of a Catboost \cite{CatBoostUnbiased} model, leading to improved model performance. In BPMN model extraction, POS tagging provides valuable context. 

\textbf{Named Entity Recognition (NER)} identifies and categorizes tokens or spans of tokens into named entities specific to the domain \cite{NER_DeepLearning}. These entities, called \textit{mentions} \cite{Neuberger}, are classified into domain-relevant types such as \textit{Actor, Activity, or Activity Data} within the BPMN context \cite{PET,Neuberger}. This classification task, based on the works in \cite{NER_DeepLearning}, was formalized in \cite{Neuberger}. 
Conditional Random Fields (CRFs) \cite{CRF_Source} are often used to predict label sequences by modelling the conditional probability of outcomes based on input sequences using log-linear distributions. 
Besides, another often used model for NER is the BertForTokenClassification model from the Hugging Face Transformers library \cite{FineTuningBERT}. 

\textbf{Relation Extraction (RE)} is a next step in IE. RE focuses on identifying and extracting relationships between entities mentioned in the text. This process involves analyzing the contextual information surrounding entities to infer meaningful connections and associations. In the context of process extraction, RE involves classifying semantic relationships that can exist between various entities, such as a \textit{Sequence flow} representing a relation between \textit{Activity} entities. 

CatBoost \cite{CatBoostUnbiased}, a widely-used model for RE \cite{Neuberger}, excels in categorical data matching and mitigates overfitting through gradient boosting on decision trees. 
When working with a high number of of categorical values, CatBoost simplifies the process and maintains accuracy without the complexities of encoding. This capability is particularly valuable in the context of BPMN model extraction, where entities such as activities, actors, and gateways need to be linked to each other through classification. For example, identifying a relationship between two activities (e.g. \textit{submit request} and \textit{review request} is essential for constructing the sequence of tasks in a business process. 

\textbf{Entity resolution (ER)} involves identifying and linking all mentions of the same entity within a text \cite{Neuberger}. The study in \cite{GrishmanInfoExtraction} illustrates this through coreference resolution, where the system identifies multiple references to the same individual, such as full names or pronouns, and associates them with the most complete mention.

\textbf{BPM Diagram extraction}, in recent years, has mostly been done by machine learning models. For training and refining machine learning models aimed at process extraction tasks, the PET dataset was introduced in \cite{PET}. It was initially developed from a corpus of textual process descriptions used by \cite{friedrich2010automated} for the extraction of process models. Additional information regarding label distributions and occurrences within the PET dataset is detailed in Section \ref{Data_Preparation:_PET_dataset_additions}. Accompanying the PET dataset, three baselines were established to assess its utility in process model extraction \cite{PET}. Baseline 1 (B1) uses Conditional Random Fields (CRF) \cite{CRF} with the IOB tagging scheme \cite{IOB_paper} to identify entities like activities and actors, categorizing it as a machine learning technique. In contrast, Baselines 2 (B2) and 3 (B3) employ rule-based strategies. B2 uses gold standard annotations to identify relationships between entities, while B3 extends B1's entity annotations with additional rules to detect relations. These baselines demonstrate the effectiveness of both rule-based and machine learning approaches in evaluating the PET dataset's efficacy for process model extraction \cite{PET}.

\section{BPMN Model Extraction from Text}\label{sec: Methodology}

This section details the methodology used. In Section \ref{subsec: generalpipeline}, the general approach to automatically extract BPMN models from text is given. Section \ref{Data_Preparation:_PET_dataset_additions} describes the datasets utilized for training the final models including our additions.

\subsection{The General Pipeline}\label{subsec: generalpipeline}
\begin{figure}
     \vspace{-15pt}
     \centering

     \includegraphics[width=0.95\textwidth]{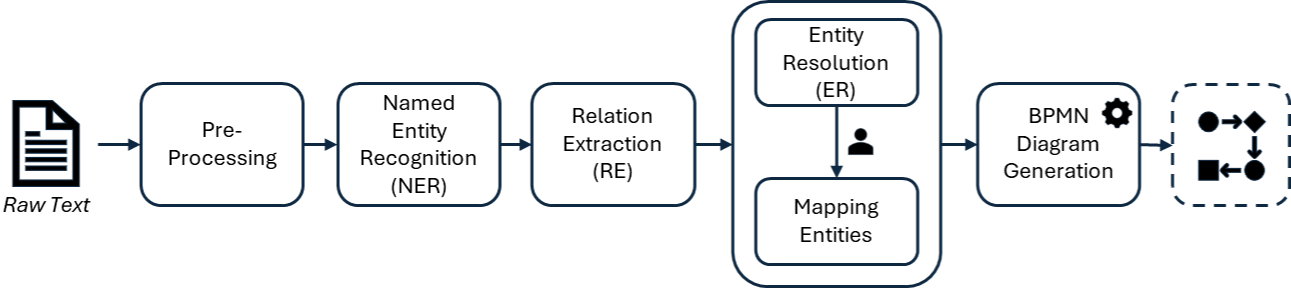}
     \caption{General Extraction pipeline}
     \label{fig:Extraction pipeline}
     \vspace{-25pt}
 \end{figure}

In Figure \ref{fig:Extraction pipeline}, the steps of our pipeline transitioning from initial text pre-processing to the final generation of BPMN diagrams are shown. They are discussed in detail below. 

\begin{enumerate}
    \item \textbf{Pre-Processing} - Process descriptions are tokenized and segmented into sentences. Segmentation allows for sentence IDs, token distances, and prevents truncation. Special characters like \texttt{["'", "-", "’", "(", "\&", ")"]} are removed to avoid labeling errors.
    \item \textbf{Named Entity Recognition} - A trained model identifies entity mentions based on a predefined BPMN tagset (see Section \ref{Data_Preparation:_PET_dataset_additions}) or assigns an \textit{"O"} label for non-entities. Labels follow the IOB format, ensuring consistency with \cite{PET,Neuberger}. 
    \item \textbf{Relation Extraction} - This phase predicts relationships between extracted mentions, assigning BPMN tagset edge labels or \textit{no\_relation}. These labels align with \cite{PET,Neuberger} and define entity interactions in the BPMN model. To enhance the model's understanding of mention relations, the training data was augmented with additional features. The data was organized into a dataframe, where each row represents a mention pair within a document, detailed by various features. These include attributes from both source and target mentions: token, type, POS tag, sentence and token ID, previous and next tags, token and sentence distances, and dependency tags
    \item \textbf{Entity Resolution} - ER is conducted after NER and RE to resolve relevant entities and minimize error propagation \cite{Neuberger}. In this phase, coreference resolution is used to group mentions into clusters, ensuring that entities are consistently represented throughout the document. ER then determines whether these mentions should be depicted as single or multiple elements in the BPMN diagram \cite{Neuberger}. 
    \item \textbf{BPMN Model Generation} - This final phase involves translating the structured overview of resolved entities and their relationships into a BPMN diagram. This step effectively visualizes the business process. 
\end{enumerate}


\subsection{The LESCHNEIDER Dataset}
\label{Data_Preparation:_PET_dataset_additions}

The original PET dataset, while extensive, has data imbalance issues including only having 8 instances of \textit{AND gateway} compared to 117 instances of \textit{XOR gateway} \cite{Neuberger}. This imbalance underscores the need for a more diverse dataset. Moreover, these challenges compromise the accuracy of predicting tags for underrepresented categories like \textit{AND Gateways} \cite{Neuberger,PET}. An overview of the original PET dataset's descriptive statistics for the mention types can be found in Table \ref{OGPET_ent_dataset_table}\footnote{This table may differ from that presented in \cite{PET} because the authors updated the dataset to version 1.1 subsequent to the publication of the paper.}.

\begin{table}[ht] 
\scriptsize
    \caption{Statistics on mention (of entity) types of PET V1.1 as presented in \cite{PET}}
    \begin{adjustbox}{center}
    \setlength{\tabcolsep}{2pt}
    \renewcommand{\arraystretch}{1.3}
    \begin{tabular}{lccccccc}
    \hline
    & \makecell{\textit{Actor}} & \makecell{\textit{Activity}} & \makecell{\textit{Activity}\\\textit{Data}} & \makecell{\textit{XOR}\\\textit{Gateway}} & \makecell{\textit{Further}\\\textit{Specification}} & \makecell{\textit{Condition}\\\textit{Specification}} & \makecell{\textit{AND}\\\textit{Gateway}} \\
    \hline
    absolute count & 449 & 502 & 459 & 117 & 64 & 80 & 8 \\
    relative count & 26.74\% & 29.90\% & 27.34\% & 6.97\% & 3.81\% & 4.76\% & 0.48\% \\
    per document & 9.98 & 11.16 & 10.20 & 2.60 & 1.42 & 1.78 & 0.18 \\
    per sentence & 1.08 & 1.20 & 1.10 & 0.28 & 0.15 & 0.19 & 0.02 \\
    \hline
    \end{tabular}
    \end{adjustbox}    \label{OGPET_ent_dataset_table}
\end{table} 

\textbf{The LESCHNEIDER dataset} is introduced in this paper as an extended version of the PET dataset and is created by drafting BPMN diagrams from a variety of real-world sources, including business articles, educational materials, and workplace procedures\footnote{\url{https://uk.indeed.com/career-advice/career-development/business-processes-examples},\url{https://quixy.com/blog/business-process/}, \url{https://www.cmwlab.com/blog/6-business-process-examples-automation-ideas/}}, all guided by established guidelines in \cite{ProcessModelingGuidelines}. These diagrams were then transformed into textual descriptions, which were annotated for NER using the IOB2 format in Excel and manually for RE, following PET standards. The new descriptive statistics can be found in Table \ref{COMBINED_PET_LESCHNEIDER_DATA_MENTION_TYPES} for the mention types.

The introduced dataset better handles the class imbalance by enhancing the dataset with 15 annotated process descriptions totaling 91 sentences. The LESCHNEIDER dataset introduces 32 new \textit{AND Gateways}, compared to 8 in the original PET dataset, by focusing on parallel process structures using phrases like ``\textit{and simultaneously}'' to identify them, similar to the studies in \cite{FriedrichBPMN,Honkisz2018} that employed similar conditional markers for detecting AND Gateways. Building on this, we use comparable markers to enhance our dataset for training machine learning models, moving away from the rule-based methods of those studies. 

In addition to featuring more parallelism, the dataset incorporates non-essential information to assess the models' capability to distinguish between essential and non-essential details. This approach addresses the issue of noise, as identified in \cite{FriedrichBPMN}, evaluating the robustness of the models under realistic conditions.

\begin{table}
\scriptsize
    \vspace{-5pt}
    \caption{Statistics on mention (of entity) types of the LESCHNEIDER dataset}
    \begin{adjustbox}{center}
    \begin{tabular}{lccccccc}
    \hline
    & \makecell{\textit{Actor}} & \makecell{\textit{Activity}} & \makecell{\textit{Activity}\\\textit{Data}} & \makecell{\textit{XOR}\\\textit{Gateway}} & \makecell{\textit{Further}\\\textit{Specification}} & \makecell{\textit{Condition}\\\textit{Specification}} & \makecell{\textit{AND}\\\textit{Gateway}} \\
    \hline
    absolute count & 542 & 613 & 568 & 136 & 86 & 98 & 40 \\
    relative count & 26.02\% & 29.43\% & 27.27\% & 6.53\% & 4.13\% & 4.70\% & 1.92\% \\
    per document & 9.03 & 10.22 & 9.47 & 2.27 & 1.43 & 1.63 & 0.67 \\
    per sentence & 1.07 & 1.21 & 1.12 & 0.27 & 0.17 & 0.19 & 0.08 \\
    \hline
    \end{tabular}
    \end{adjustbox}    \label{COMBINED_PET_LESCHNEIDER_DATA_MENTION_TYPES}
    \vspace{-5pt}
\end{table} 

In this study, \textit{gold standard data} refers to documents in the combined dataset that are already annotated with correct tags. These annotations serve as the correctly labeled dataset against which the performance of our natural language processing models is validated. This gold standard data is crucial, as it provides the definitive tagging of descriptions necessary for ensuring the integrity of our NLP applications.

\section{Experimental Evaluation}\label{sec: evaluation}

In this section, the research questions are provided, followed by the rationale behind choosing certain NLP techniques or models for each step in the pipeline as discussed in \ref{subsec: generalpipeline} which includes the choice for the hyperoptimization settings and implementation details of various setups are provided. All models were trained on the L4 GPU in Google Colab. 

\subsection{Research Questions}\label{sec: RQs}

The research questions answered in this paper are designed to address the identified gaps and enhance the efficacy of process model extraction:

\begin{questions}
\item Is the PET dataset diverse enough to train models capable of effectively managing various writing styles and process domains for BPMN extraction?

\item Compared to the original PET dataset, are models trained on the LESCHNEIDER dataset better capable of accurately identifying parallelism (AND Gateways) in unseen texts?

\item Among advanced pre-trained models RoBERTa and BERT, which demonstrate greater effectiveness in accurately extracting BPMN from textual descriptions with varied writing styles and domains?
\end{questions}

To answer these research questions, we conduct three key experiments. The first experiment evaluates two methods for NER using the PET dataset and LESCHNEIDER dataset. The first approach tests the CRF model, originally used in the PET study, by performing a five-fold cross-validation. The second approach involves training models on the PET dataset and testing them on the LESCHNEIDER dataset, thereby assessing both the dataset’s diversity (RQ1) and the ability of LESCHNEIDER-trained models to improve parallelism detection (RQ2). The second experiment performs a five-fold cross-validation of selected models for both the NER and RE modules using a combined dataset. This experiment specifically addresses RQ3 by comparing the performance of RoBERTa and BERT in extracting BPMN elements from varied textual descriptions. The third experiment evaluates the full pipeline by integrating the best fine-tuned model for NER with a pre-trained CatBoost model for relation extraction. The evaluation is conducted both with and without coreference resolution to assess its impact on overall extraction performance.

\subsection{Model Selection per NLP subtask}\label{sec: modelselection}

For the NER models of our pipeline, the models considered include CRF, BertForTokenClassification,
and RobertaForTokenClassification. For the CRP, the features include token attributes (text, POS tags, suffixes, capitalization) and GloVe embeddings for semantic context \cite{GloVe}. This setup allows CRFs to accurately label tokens, establishing a baseline for comparison with other methods \cite{PET,Neuberger}. Next, we fine-tuned four pre-trained BERT models (\texttt{bert-base-uncased}, \texttt{bert-base-cased}, \texttt{bert-large-uncased},~and \texttt{bert\allowbreak-large-cased}) on our dataset for BPMN entity extraction. Despite BERT's tokenizer, we pre-processed the text to remove special characters. To compare other LLMs on our NER task, we evaluated the Robustly Optimized BERT Pretraining Approach \textbf{RoBERTa} model \cite{RoBERTa}. RoBERTa improves on BERT by extending training, using longer sequences, larger batch sizes, and eliminating the Next Sentence Prediction task \cite{RoBERTa}. It is trained on a larger dataset with dynamic masking to enhance context understanding \cite{RoBERTa}. We fine-tuned \textbf{RobertaForTokenClassification} on our dataset using two pre-trained models: \texttt{roberta-base}\footnote{\url{https://huggingface.co/FacebookAI/roberta-base}} and \texttt{roberta-large}\footnote{\url{https://huggingface.co/FacebookAI/roberta-large}}. Both models are case-sensitive and share the same architecture as their BERT counterparts. 

For the RE, A CatBoost model was trained to classify relationships using a comprehensive feature set for each bidirectional pair of mentions. To address the imbalance in the relation dataset, data sampling is employed for more accurate class predictions. The \textit{flow} class is prioritized because it fundamentally defines the sequential interaction of activities within process models, directly influencing the logical progression of tasks. Enhancing the performance for the \textit{flow} class, while maintaining accuracy for other classes, requires targeted data sampling techniques. The term \textit{baseline} in this section refers to the combined unsampled dataset from the PET and LESCHNEIDER datasets, as detailed in \ref{Data_Preparation:_PET_dataset_additions}.
After feature extraction, we examined three sampling techniques to address the challenges posed by data imbalance: \emph{negative sampling} \cite{Neuberger}, \emph{Synthetic Minority Over-sampling Technique (SMOTE)}\cite{SMOTE_Original} and \emph{Random Over Sampling (ROS)}. For the negative sampling, to refine the model's accuracy, we experimented with various negative sampling rates, focusing on their impact on precision and recall. We applied SMOTE on both negative sampled and baseline data, specifically targeting the \textit{flow} relation to double in occurrence. Finally, we applied ROS to the \textit{flow} relation on both the baseline data and the negative sampled data to be twice the current \textit{flow} count. 

To maintain clarity, our BPMN diagrams will focus on core BPMN elements supported by the NER and RE tagset, excluding less common elements like sub-processes and special events \cite{Howmuchlanguageisenough_core_BPMNelements}. Actors will be depicted within ellipses and connected to activities with colored edges, simplifying the diagrams for better interpretability. Although our diagram generation supports loops, we intentionally avoided annotating them in our dataset to simplify the process and focus on parallelism. In addition, the PET dataset \cite{PET} itself does not explicitly focus on loops, aligning with our streamlined approach.
We employ the Directed Graph (DiGraph) for generating BPMN diagrams. This tool facilitates the creation of diagrams in \texttt{.dot} file format, allowing for customization and user refinement. Our implementation of DiGraph includes a feature to strategically place unconnected entities at the top left of the diagram to ensure all extracted elements are visible. 

\subsection{Implementation Details}

For the BERT and RoBERTa models, we used default hyperparameters except for learning rate (LR), batch size and epochs following \cite{FineTuningBERT} to avoid catastrophic forgetting related to LR. We tested LRs of \texttt{[2e-5, 3e-5, 4e-5, 5e-5]} following \cite{BERT} and selected optimal rates based on F1 scores. A smaller batch size of eight led to quicker convergence and better performance on complex labels like \textit{AND Gateway}. 
The number of epochs was set empirically by averaging optimal epochs across folds and adding two, to ensure comprehensive learning without overfitting, monitored via validation and training loss. The CRF model used baseline hyperparameters from \cite{PET}. The code implementation and datasets can be found in the \href{https://github.com/AlexanderPaulStevens/BPMN-Model-Generation-from-Text}{GitHub repository}.
\section{Results}\label{sec: Discussion}

This section evaluates the selected models and methodologies. 

\subsection{Performance metrics}
\label{Evaluation_Performance_metrics}

The NER and RE modules are evaluated using Precision $\text{Precision} = \frac{TP}{TP + FP}$, Recall $\text{Recall} = \frac{TP}{TP + FN}$, and F1-score $\text{F1} = 2 \cdot \frac{\text{Precision} \cdot \text{Recall}}{\text{Precision} + \text{Recall}}$:
These metrics are crucial for accurate entity identification, preventing error propagation that affects precision and recall. The F1-score is particularly useful for our imbalanced dataset, balancing precision and recall. For multi-label tasks, we use three averaging methods; Micro F1, Macro F1, and Weighted F1.

\subsection{Experiment 1: Baseline CRF model performance on PET vs LESCHNEIDER dataset}
\label{results_experiment1}

This experiment addresses \textbf{RQ1} and \textbf{RQ2} by examining whether the PET dataset can train models to generalize across writing styles and whether adding the LESCHNEIDER dataset, which includes \textit{AND Gateways}, improves model performance.

We first evaluated a CRF model trained solely on PET using 5-fold cross-validation as a baseline, with results shown in Table \ref{tab:CRF_Model_results} under ``\textbf{(a)Baseline CRF}''. The Baseline CRF model performed well, achieving a weighted average F1 score of 0.72, excelling in categories like \textit{B-Actor} (F1 = 0.80), \textit{B-Activity} (F1 = 0.78), and \textit{B-XOR Gateway} (F1 = 0.78), where labels typically have clear identifiers \cite{Neuberger}.

Next, we trained the model on PET and tested it on LESCHNEIDER. As seen in column "\textbf{(b)CRF Model 2}" of Table \ref{tab:CRF_Model_results}, performance declined, with the weighted average F1 score dropping from 0.72 to 0.61, indicating possible challenges in generalizing to different writing styles. The decline in performance for CRF Model 2 was primarily driven by significant drops in precision, recall, and F1 scores for labels such as \textit{B-Activity} (F1 score from 0.78 to 0.67), \textit{I-Activity} (F1 score from 0.41 to 0.20), and \textit{B-Further Specification} (F1 score from 0.30 to 0.13). Additionally, the \textit{I-XOR Gateway} label saw a complete drop to zero in all metrics. 

\begin{table}[ht]
\scriptsize
    \centering
    \caption{Overview of CRF models results on different datasets: (a) trained \& tested on PET, (b) trained on PET \& tested on LESCHNEIDER, (c) trained \& tested on combined dataset.}
    \label{tab:CRF_Model_results}
    \setlength{\tabcolsep}{2pt}
    \begin{tabular}{lccc|ccc|ccc}
        & \multicolumn{3}{c|}{\textbf{(a)Baseline CRF}} & \multicolumn{3}{c|}{\textbf{(b)CRF Model 2}} & \multicolumn{3}{c}{\textbf{(c)CRF Model 3}} \\
        \hline
        & Prec. & Recall & F1 & Prec. & Recall & F1 & Prec. & Recall & F1 \\
        \hline
        B-AND Gateway & 0.00 & 0.00 & 0.00 & 0.00 & 0.00 & 0.00 & 0.50 & 0.15 & 0.23 \\
        I-AND Gateway & 0.00 & 0.00 & 0.00 & 0.00 & 0.00 & 0.00 & 0.07 & 0.06 & 0.07 \\
        B-Activity    & 0.81 & 0.76 & 0.78 & 0.80 & 0.58 & 0.67 & 0.80 & 0.77 & 0.78 \\
        I-Activity    & 0.53 & 0.33 & 0.41 & 1.00 & 0.11 & 0.20 & 0.67 & 0.38 & 0.48 \\
        B-Activity Data & 0.76 & 0.71 & 0.73 & 0.79 & 0.52 & 0.63 & 0.75 & 0.70 & 0.72 \\
        I-Activity Data & 0.69 & 0.70 & 0.69 & 0.71 & 0.56 & 0.63 & 0.70 & 0.71 & 0.70 \\
        B-Actor        & 0.81 & 0.78 & 0.80 & 0.95 & 0.63 & 0.76 & 0.82 & 0.81 & 0.81 \\
        I-Actor        & 0.77 & 0.77 & 0.77 & 0.97 & 0.66 & 0.79 & 0.78 & 0.78 & 0.78 \\
        B-Condition Spec. & 0.87 & 0.65 & 0.74 & 0.88 & 0.78 & 0.82 & 0.88 & 0.69 & 0.78 \\
        I-Condition Spec. & 0.74 & 0.59 & 0.65 & 0.95 & 0.86 & 0.90 & 0.78 & 0.66 & 0.71 \\
        B-Further Spec. & 0.42 & 0.23 & 0.30 & 0.22 & 0.09 & 0.13 & 0.49 & 0.27 & 0.35 \\
        I-Further Spec. & 0.27 & 0.19 & 0.22 & 0.28 & 0.24 & 0.26 & 0.39 & 0.25 & 0.31 \\
        B-XOR Gateway  & 0.83 & 0.74 & 0.78 & 1.00 & 0.84 & 0.91 & 0.81 & 0.77 & 0.79 \\
        I-XOR Gateway  & 0.80 & 0.39 & 0.52 & 0.00 & 0.00 & 0.00 & 0.57 & 0.34 & 0.43 \\
        O              & 0.73 & 0.80 & 0.76 & 0.55 & 0.90 & 0.68 & 0.73 & 0.80 & 0.76 \\
        \hline
        \textbf{Weighted Average} & 0.72 & 0.72 & 0.72 & 0.63 & 0.64 & 0.61 & 0.72 & 0.73 & 0.72 \\
    \end{tabular}
    \vspace{-15pt}
\end{table}

Finally, we evaluated a CRF model on the combined dataset with 5-fold cross-validation, shown in ``\textbf{(c)CRF Model 3}'' of Table \ref{tab:CRF_Model_results}, and we can see improved performance across most tags compared to the PET-only trained model. The inclusion of LESCHNEIDER improved performance across most tags, particularly on \textit{AND gateways}, though no improvements were seen for \textit{O}, \textit{B-Activity Data}, \textit{I-Activity Data}, \textit{B-Actor}, \textit{I-Actor}, \textit{B-XOR Gateway}, and \textit{B-Activity}. Overall, the enhanced dataset led to performance gains for 7 out of 15 tags, with only \textit{I-XOR Gateway} showing a decline, indicating the positive impact of the added data on model accuracy.

\subsection{Experiment 2: Comparative analysis of NER and RE models} 

For the second experiment, we performed a 5-fold cross-validation on the models selected in Section \ref{sec: Methodology}. The cross-validation was performed on the complete dataset with PET \cite{PET} and LESCHNEIDER combined as described in Section \ref{Data_Preparation:_PET_dataset_additions}.

\begin{figure}[ht]
    \centering    
    \begin{subfigure}[b]{0.49\linewidth}
        \centering
        \includegraphics[width=\linewidth]{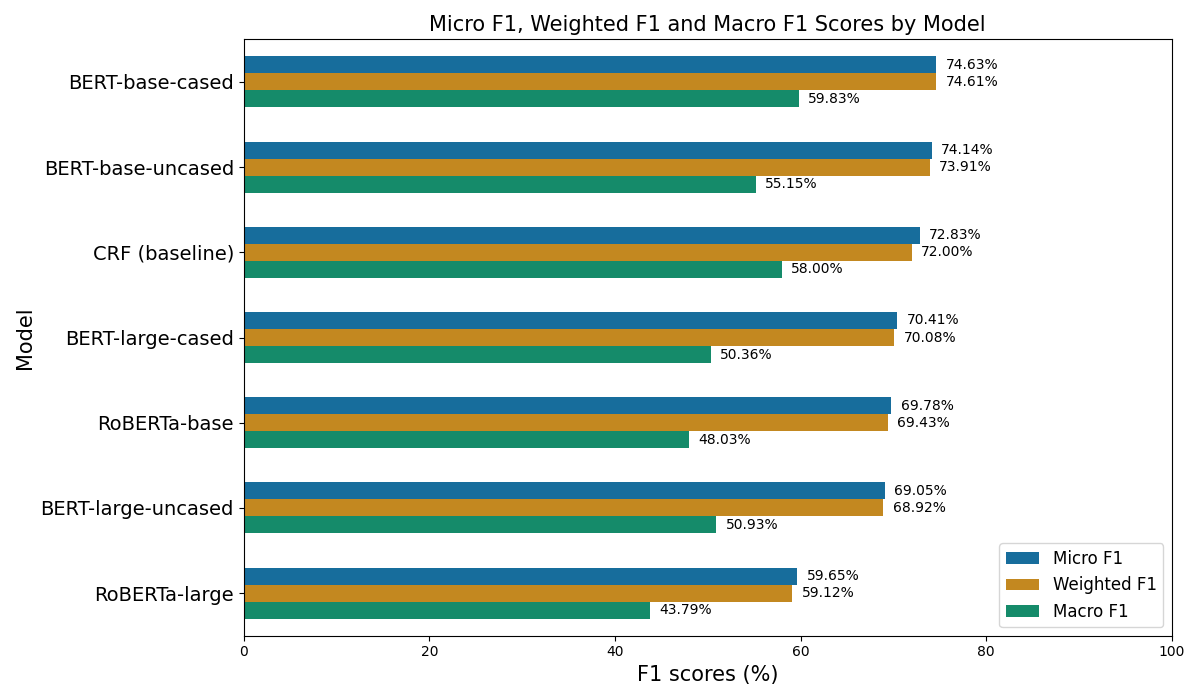}
        \caption{Overall 5-fold cross-validation results for NER per model}
        \label{fig:subfig_NER_results_per_model}
    \end{subfigure}
    \begin{subfigure}[b]{0.49\linewidth}
        \centering
        \includegraphics[width=\linewidth]{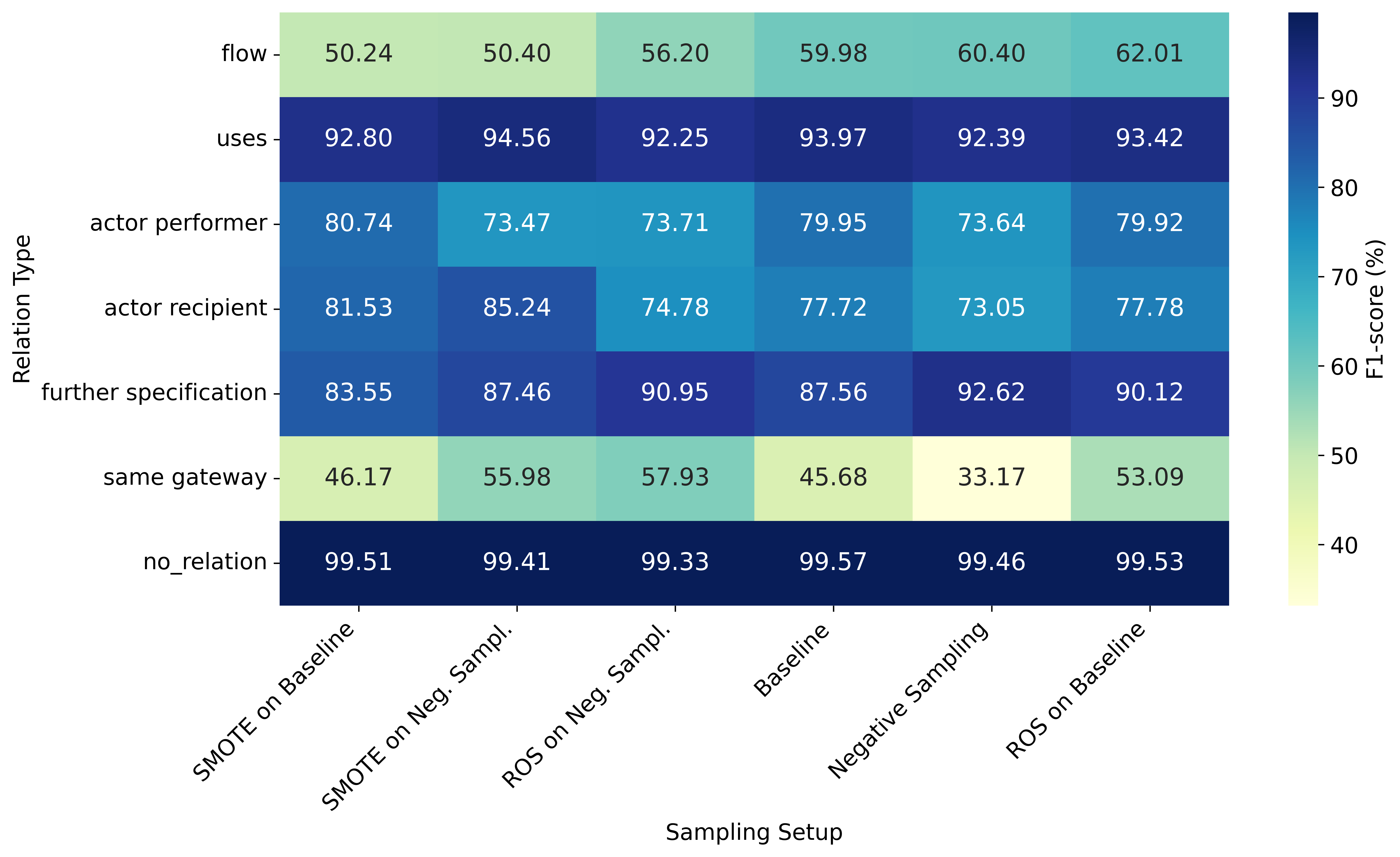}
        \caption{F1 Scores (\%) of relation types across different sampling setups}
        \label{fig:subfig_heatmap_sampling_performance}
    \end{subfigure}
    
    \caption{Model performance visualizations: (a) NER results, (b) Relation type F1 scores}
    \label{fig:combined_model_results}
\end{figure}

For the NER module of this experiment, the individual model performances are visualized and compared in Figure \ref{fig:subfig_NER_results_per_model}, which displays the metrics averaged over all five folds. The BERT-base-cased model outperforms others, achieving the highest F1-scores across all types, indicating strong and balanced performance. The BERT-base-uncased model follows closely, but its slightly lower performance highlights the importance of case information for entity extraction.

For relation extraction, we compared three sampling techniques: Negative Sampling, SMOTE, and ROS (as detailed in Section \ref{sec: Methodology}), using 5-fold cross-validation. The result of this comparison can be seen in Figure \ref{fig:subfig_heatmap_sampling_performance}. ROS showed the best performance for the \textit{flow} class while also maintaining strong results across other categories. As a result, ROS was selected for training our final CatBoost model in the pipeline.

\subsection{Experiment 3: Pipeline evaluation} 

We assess the interpretability and accuracy of BPMN diagrams generated by our pipeline using both quantitative and qualitative methods, focusing on clarity and utility. We processed six test documents through the pipeline to generate diagrams, which are then compared to gold standard equivalents. 

The evaluation counts elements in the diagram: \#GOLD (gold standard elements), \#PREDICTED (pipeline predictions), and \#CORRECT (accurately predicted elements). Precision (\(\text{P} = \frac{\#CORRECT}{\#PREDICTED}\)) measures the proportion of correctly predicted elements, recall (\(\text{R} = \frac{\#CORRECT}{\#GOLD}\)) represents the proportion of correctly identified elements from the gold standard, and F1-score (\(\text{F1} = 2 \cdot \frac{\text{P} \cdot \text{R}}{\text{P} + \text{R}}\)) is their harmonic mean. Table \ref{PIPELINE_EVALUATION_AGGREGATED} presents the aggregated results.

The pipeline showed strong performance on the \textit{elements} with a aggregated F1 of 89.2\% whilst the performance of the \textit{relations} scored lower with a total F1 of 73.7\% with more variability across the testing documents.
\vspace{-10mm}
\begin{table}[h]
 \scriptsize
    \caption{Pipeline Performance for Elements and Relations}
    \begin{adjustbox}{center}
    \renewcommand{\arraystretch}{1.2}
    \begin{tabular}{lccccccc}
    \hline
    & \#Words & \#Gold & \#Pred & \#Correct & Precision & Recall & F1 \\
    \hline
    \multicolumn{8}{c}{\textbf{Elements}} \\
    \hline
    Doc 1  & 104  & 21  & 24  & 20  & 83.3\%  & 95.2\%  & 88.9\%  \\
    Doc 2  & 69   & 17  & 18  & 17  & 94.4\%  & 100.0\% & 97.1\%  \\
    Doc 3  & 51   & 12  & 12  & 12  & 100.0\% & 100.0\% & 100.0\% \\
    Doc 4  & 36   & 13  & 8   & 8   & 100.0\% & 61.5\%  & 76.2\%  \\
    Doc 5  & 88   & 19  & 18  & 15  & 83.3\%  & 78.9\%  & 81.1\%  \\
    Doc 6  & 81   & 21  & 21  & 19  & 90.5\%  & 90.5\%  & 90.5\%  \\
    \hline
    Total  & 429  & 103 & 101 & 91  & 90.1\%  & 88.3\%  & 89.2\%  \\
    \hline
    \multicolumn{8}{c}{\textbf{Relations (Edges)}} \\
    \hline
    Doc 1  & 104  & 35  & 37  & 19  & 51.4\%  & 54.3\%  & 52.8\%  \\
    Doc 2  & 69   & 32  & 31  & 29  & 93.5\%  & 90.6\%  & 92.1\%  \\
    Doc 3  & 51   & 20  & 20  & 19  & 95.0\%  & 95.0\%  & 95.0\%  \\
    Doc 4  & 36   & 23  & 15  & 11  & 73.3\%  & 47.8\%  & 57.9\%  \\
    Doc 5  & 88   & 33  & 31  & 23  & 74.2\%  & 69.7\%  & 71.9\%  \\
    Doc 6  & 81   & 29  & 28  & 22  & 78.6\%  & 75.9\%  & 77.2\%  \\
    \hline
    Total  & 429  & 172 & 162 & 123 & 75.9\%  & 71.5\%  & 73.7\%  \\
    \hline
    \end{tabular}
    \end{adjustbox}
    \label{PIPELINE_EVALUATION_AGGREGATED}
\end{table}
\vspace{-10mm}

\subsection{Discussion}
While the automated pipeline shows promise, challenges and limitations remain.
First, in addressing \textbf{RQ1}, our evaluation of the PET dataset shows that it effectively trains models to extract basic BPMN elements like \textit{B-Actor} and \textit{B-XOR Gateway}, thanks to its diverse and representative samples of these labels. However, it faces challenges with more complex labels such as \textit{I-XOR Gateway}, often linked to the term ``case'' in the dataset, and \textit{AND Gateways}, impacting the model's ability to manage parallel processes. Moreover, the dataset performs poorly in identifying ``\textit{O}'' labels when applied to another dataset, indicating a lack of non-essential examples that are needed to discern important details. These issues limit the dataset’s generalizability and its capacity to handle complex structures and varied writing styles effectively.
Second, in response to \textbf{RQ2}, we mitigated some limitations of the PET dataset by integrating the LESCHNEIDER dataset, which significantly improved the model's ability to detect parallel structures. This integration resulted in a notable rise in the F1 score for \textit{B-AND Gateway} from 0\% to 23\% and an increase in precision to 50\%. Most labels either improved or maintained their previous performance levels, but challenges persist with complex labels like the \textit{I-XOR Gateway}, linked to specific phrases in the PET dataset that do not generalize well. Despite these improvements, the model's capability to accurately represent complex BPMN structures, such as parallel activity structures, remains limited, emphasizing the need for further dataset and model refinements to enhance overall performance.
Finally, in addressing \textbf{RQ3}, the BERT-base-cased model demonstrates superior performance in recognizing frequent and simpler labels, also achieving the highest F1 score for \textit{B-AND Gateway}. In contrast, larger models such as BERT-large and RoBERTa-large tend to overfit when trained on the relatively small dataset used in our study, leading to diminished effectiveness for these models. 

\section{Conclusion}\label{sec: Conclusion}

This work presents an automated pipeline for BPMN model generation, leveraging NLP techniques and machine learning models such as CRF, BERT, and RoBERTa, to extract process elements from textual descriptions. The models were evaluated on an augmented PET dataset, enriched with additional parallel gateways and noise elements to improve robustness.

Key contributions include enhancing the PET dataset to improve parallelism detection and filter out irrelevant data, leading to improved model performance, especially for \textit{AND Gateways}. 
However, training over PET struggled with generalizing to diverse writing styles due to its limited vocabulary for certain elements. A comparative analysis of BERT and RoBERTa showed that \texttt{BERT-base-cased} performed best, likely due to its balanced complexity given the dataset’s size, whereas larger models risked overfitting. 

A key limitation is the CatBoost model’s low F1 score (62\%) for \textit{flow} relations, which hinders accurate BPMN diagram generation and the models struggle with implicit gateway closures.

Future work includes exploring LLMs for relation extraction, hybrid rule-based and ML approaches for better generalization, and expanding datasets to capture diverse language styles. Automated error-handling mechanisms, such as ChatGPT-based validation, could further enhance BPMN generation efficiency.

%
%
%
%

\end{document}